\documentclass{IEEEcsmag}

\usepackage[colorlinks,urlcolor=blue,linkcolor=blue,citecolor=blue]{hyperref}

\usepackage{upmath}
\usepackage{multirow}
\usepackage{graphicx}
\usepackage{subfigure}
\usepackage{todonotes}
\usepackage{amssymb}

\begin{document}


\title{Blockchain meets Biometrics: \LARGE{Concepts, Application to Template Protection, and Trends}}

\author{Oscar Delgado-Mohatar, Julian Fierrez, Ruben Tolosana and Ruben Vera-Rodriguez}
\affil{School of Engineering, Universidad Autonoma de Madrid, Madrid, Spain}

%
%
%


\begin{abstract}
Blockchain technologies provide excellent architectures and practical tools for securing and managing the sensitive and private data stored in biometric templates, but at a cost. We discuss opportunities and challenges in the integration of blockchain and biometrics, with emphasis in biometric template storage and protection, a key problem in biometrics still largely unsolved. Key tradeoffs involved in that integration, namely, latency, processing time, economic cost, and biometric performance are experimentally studied through the implementation of a smart contract on the Ethereum blockchain platform, which is publicly available in github for research purposes.
\end{abstract}

\maketitle

\chapterinitial{Blockchain and biometrics}\ are two of the most disruptive technologies of this era.\footnote{\url{https://github.com/BiDAlab/BlockchainBiometrics}}

Blockchain technology provides an immutable and decentralized data registry, optionally with the capability of executing distributed code in a secure way. Its origins are linked to Bitcoin cryptocurrency (2009), solving an old opened problem in the cryptographic community since the 80's: the design of a distributed algorithm of consensus on economic transactions without the participation and control of a central authority. Nevertheless, nothing prevents any other digital data from being stored in the blockchain instead of just economic transactions. This aspect opens the doors to many different potential applications such as smart energy and grids~\cite{Magnani2018}, healthcare~\cite{GORDON2018224}, and smart devices or digital identity schemes~\cite{buchmann2017enhancing}, among others.

Biometrics could be one of these potential applications to make the most of blockchain technology. From its origins in the early 60's up to now, a significant progress has been achieved, incorporating them in most daily devices such as smartphones. Biometrics refers to the automated recognition of individuals through the use of physiological (e.g., face, fingerprint) or behavioral (e.g., voice, handwritten signature) traits~\cite{jain_2016}. Its main advantages over traditional authentication methods are: \textit{i)} no need to carry tokens or remember passwords, \textit{ii)} they are harder to circumvent, and \textit{iii)} provide at the same time a stronger link between the subject and the action or event.

\begin{table*}[t]
\centering
\caption{Blockchain technology taxonomy. Consensus algorithms, PoW (proof-of-work), PoS (proof-of-stake), PoA (proof-of-authority), PoI (proof-of-importance), BFT (Byzantine Fault Tolerance). $^\star$NEM can be also configured as a public blockchain.}
\label{tab:blockchain-types}
\resizebox{\textwidth}{!}{%
\begin{tabular}{|c|l|l|l|c|c|c|c|c|}
\hline
\textbf{\begin{tabular}[c]{@{}c@{}}Type\\ (Read access)\end{tabular}} & \multicolumn{1}{c|}{\textbf{\begin{tabular}[c]{@{}c@{}}Access control\\ (Write access)\end{tabular}}} & \multicolumn{1}{c|}{\textbf{Main platforms}} & \multicolumn{1}{c|}{\textbf{\begin{tabular}[c]{@{}c@{}}Consensus \\ algorithm\end{tabular}}} & \textbf{Participants} & \textbf{\begin{tabular}[c]{@{}c@{}}Data \\ privacy\end{tabular}} & \textbf{\begin{tabular}[c]{@{}c@{}}Smart contract\\ capability\end{tabular}} & \textbf{\begin{tabular}[c]{@{}c@{}}Transaction\\ confirmation\\ time\end{tabular}} & \textbf{\begin{tabular}[c]{@{}c@{}}Associated\\ cryptocurrency\end{tabular}} \\ \hline
\multirow{4}{*}{Public} & \multirow{2}{*}{Permissionless} & Bitcoin & Proof-of-work & \multirow{4}{*}{Anonymous} & No & No & \multirow{4}{*}{\begin{tabular}[c]{@{}c@{}}Long\\ (minutes)\end{tabular}} & BTC \\ \cline{3-4} \cline{6-7} \cline{9-9}
 &  & Ethereum & \begin{tabular}[c]{@{}l@{}}PoW (now)\\ PoS (after Casper)\end{tabular} &  & No & Yes &  & ETH \\ \cline{2-4} \cline{6-7} \cline{9-9}
 & \multirow{2}{*}{Permissioned} & Quorum  & Flexible (Raft, BFT, PoA) &  & Optional & Yes &  & - \\ \cline{3-4} \cline{6-7} \cline{9-9}
 &  & Ripple  & XLCP (Byzantine based) &  & No & No &  & XRP \\ \hline
\multirow{4}{*}{Private} & \multirow{2}{*}{Permissionless} & LTO & PoI & \multirow{4}{*}{\begin{tabular}[c]{@{}c@{}}Identified, \\ \\ trusted\end{tabular}} & Yes & Yes & \multirow{4}{*}{\begin{tabular}[c]{@{}c@{}}Short\\ (seconds)\end{tabular}} & - \\ \cline{3-4} \cline{6-7} \cline{9-9}
 &  & Parity (Ethereum) & PoA &  & Yes & Yes &  & - \\ \cline{2-4} \cline{6-7} \cline{9-9}
 & \multirow{2}{*}{Permissioned} & Hyperledger  & Flexible (PBFT) &  & Yes & Yes &  & - \\ \cline{3-4} \cline{6-7} \cline{9-9}
 &  & NEM$^\star$ & PoI &  & Yes & Yes &  & - \\ \hline
\end{tabular}%
}
\end{table*}

However, and despite the advantages and potential of each of them, blockchain and biometrics have been largely developed independently so far. Some questions arise in this regard: what benefits could achieve the combination of them? what challenges and limitations should be considered for their integration?

As a first approximation, blockchain could provide biometrics with some desirable features such as immutability, accountability, availability or universal access~\cite{nanda19block}. These properties enabled by blockchain technology may be very useful, among other applications in biometrics, to secure the biometric templates \cite{nanda15template}, and to assure privacy in biometric systems \cite{bringer13privacy}.

Blockchain technology could also take advantage of biometrics in many different ways, e.g., improving the current distributed digital identity schemes on blockchain. Another interesting application is related to smart devices. A smart device is any digital or physical asset with access to a blockchain that can perform actions and make decisions based on the information stored there. For example, a car could be fully managed (rented or bought) through a smart contract. However, an adequate identification of the user is not fully solved yet. In this case, an authentication protocol based on biometrics could significantly raise the current security level.

\section{BLOCKCHAIN TECHNOLOGY}
Blockchain is essentially a decentralized public ledger of all data and transactions that have ever been executed in the system. These transactions are recorded in blocks that are created and added to the blockchain in a linear and chronological order (immutable). But, who is in charge of the correct operation of the blockchain network? Any electronic device, including computer, phone or even a printer, as long as it is connected to the Internet and as such has an IP address. These electronic devices are known as nodes, and their main roles are: \textit{i)} maintaining a copy of the blockchain, and \textit{ii)} validating relaying transactions, and copying them in the blockchain.

Since its initial application to Bitcoin cryptocurrency, the original idea of a universal and public blockchain has greatly evolved into new architectures.

\subsection{Architectures}
Blockchain technology is not a monolithic concept any more. It can be classified according to different criteria \cite{ledger140}. The most significant classifications are related to the access/control layer and the consensus algorithms. Table~\ref{tab:blockchain-types} shows a taxonomy of the most popular blockchain technologies regarding type, access control, platform, consensus algorithm, participants, data privacy, smart contract capability, time, and associated cryptocurrency.

According to the first criterion, blockchains are usually categorized as public and private. The first type allows anyone to join the network, and read or write data and transactions. Therefore, public blockchains are usually designed with a built-in economic incentive for allowing anonymous and universal access. This incentive purpose is twofold: \textit{i)} it encourages the participation of users in the network, which is important for the consensus algorithm and provides security; and \textit{ii)} it encourages the generation of more cryptocurrency. This process uses a mining mechanism based on the resolution of computationally intensive cryptographic puzzles. The most popular blockchain platforms belong to this group, e.g., Bitcoin or Ethereum.

Private blockchains, on the other hand, impose constrains in the admission process to the network (through authentication of the nodes using X.509 certificates, for example). For this reason, private blockchains are mainly considered in applications where users must be fully identified and trusted.

In some cases, permissions for reading and writing are segregated, obtaining different combinations of public/private (for the reading access), and permissionless/permissioned blockchains (for the writing access). For example, in a public permissioned blockchain any participant could freely join the network, but there would be some kind of control over who could read or write transactions. In addition, in order to protect the privacy of transactions, the data is typically encrypted.

Finally, some authors also consider an intermediate architecture between public and private blockchains, called consortium blockchains. These architectures are permissioned, partially private and with semi-decentralized architectures, specially considered in business and financial scenarios with a small number of participants.

\subsection{Smart Contracts}
Apart from storing data, blockchain technology reaches their full potential through the so-called smart contracts. A smart contract is, essentially, a piece of code executed in a secure environment that controls digital assets. This is a very well-known concept inside the cryptographic community. Some examples of these secure environments include regular servers controlled by ``trusted parties'', decentralized networks (blockchains), or servers with secure hardware (e.g., SGX) \cite{Karande2017}.

Many public blockchains support the execution of smart contracts, but Ethereum \cite{Dannen2017} is currently considered the most reliable, secure and used. In essence, Ethereum could be seen as a distributed computer, with capability to execute programs written in Turing-complete, high-level programming languages. These programs comprise a collection of pre-defined instructions and data that have been recorded at a specific address of a blockchain. For biometric purposes, a smart contract running in a blockchain can assure a semantically correct execution.

\subsection{Challenges and Limitations}
\label{sec:privacy}
Despite the advantages and new opportunities already described, the combination of both blockchain and biometrics is not straightforward due to the limitations of the current blockchain technology. Among them, it is important to remark: \textit{i)} its transaction processing capacity is currently very low (around tens of transactions per second), \textit{ii)} its actual design implies that all system transactions must be stored, which makes the storage space necessary for its management to grow very quickly, and \textit{iii)} its robustness against different types of attacks has not been sufficiently studied yet.

We now discuss the limitations of blockchain public networks for the deployment and operation of biometric systems.

\begin{itemize}
\label{sec:coste}
\item \textbf{Economic cost of executing smart contracts:} in order to support smart contracts in blockchains (like Ethereum), and to reward the nodes that use their computing capacity to maintain the system, each instruction executed requires the payment of a fee in a cryptocurrency (called gas). Simple instructions (such as a sum) cost 1 gas, while others can cost significantly more (e.g., the calculation of a SHA3 hash costs 30 gas). Additionally, the storage space is especially expensive (around 20k gas for every 256 bits). Therefore, one of the first research problems would be minimizing the cost of running a biometric system (totally or partially) in a blockchain, which includes developing efficient ways to code biometrics modules in smart contracts.

\item \textbf{Privacy:} by design, all operations carried out in a public blockchain are known by all the participating nodes. Thus, it is not possible to directly use secret cryptographic keys, as this would reduce the number of potential applications.
Regarding privacy in public blockchains, three main layers are considered in general: \textit{i)} participants, \textit{ii)} terms, and \textit{iii)} data. The first one ensures participants to remain anonymous both inside and outside of the blockchain. This is achieved with cryptographic mechanisms like ring signatures, stealth addresses, mixing, or storage of private data off-chain. Second, privacy of terms keeps the logic of the smart contracts secret, by using range proofs or Pedersen commitments. Finally, and the most important for biometrics, the goal of the data privacy layer is to keep transactions, smart contracts, and other data such as biometric templates, encrypted at all times, both on-chain and off-chain. The cryptographic tools used include zero-knowledge proofs (ZKP) and zk-SNARKS, Pedersen commitments, or off-chain privacy layers like hardware-based trusted execution environments (TEEs). However, the application of these cryptographic tools are still very limited for blockchains. For example, Ethereum just included at the end of 2017 basic verification capabilities for ZKPs. More advanced cryptographic tools have been only developed to target special cases like Aztec~\cite{Williamson2018} or ZK range proofs~\cite{Koens2018}. In addition, it should be noted that ZKP transactions would be expensive and computationally intensive ($\sim$~1,5M gas/verification).

\item \textbf{Processing capability:} another important limitation is related to its processing capability. Ethereum, for example, is able to run just around a dozen transactions per second, what it could be not enough for some scenarios. Additionally, there is a minimum confirmation time before considering that the transaction has been properly added to the blockchain. This time can oscillate among different blockchains, from tens of seconds to minutes, reducing its usability for biometric systems.

\item \textbf{Scalability:} this is one of the main handicaps of the technology from its origin as, theoretically, all nodes of the blockchain network must store all blocks of the blockchain network. Currently, the size of the public blockchains (Bitcoin and Ethereum) is around 200GB, but this is growing very fast. This can be a problem for some application scenarios such as the Internet of Things (IoT).

\item \textbf{Security:} as novel technology, blockchain security characterization is still a work in progress. Among all possible attacks, it is worth mentioning the attack known as ``51\% attack''~\cite{Eyal2014}. If an attacker gains more than 50\% of the computational capacity of any public or private blockchain, he could reverse or falsify transactions. This attack applies even to blockchain with consensus algorithms not based in proof-of-works schemes, like PoS or PoA, typically used in private or consortium topologies. However, the main security problems suffered to date are mainly related to programming errors, e.g., the DAO attack happened in 2016, which put at risk the whole Ethereum ecosystem~\cite{Atzei2017}.
\end{itemize}

\begin{figure*}[t]
\centering
\centerline{\includegraphics[width=0.75\textwidth]{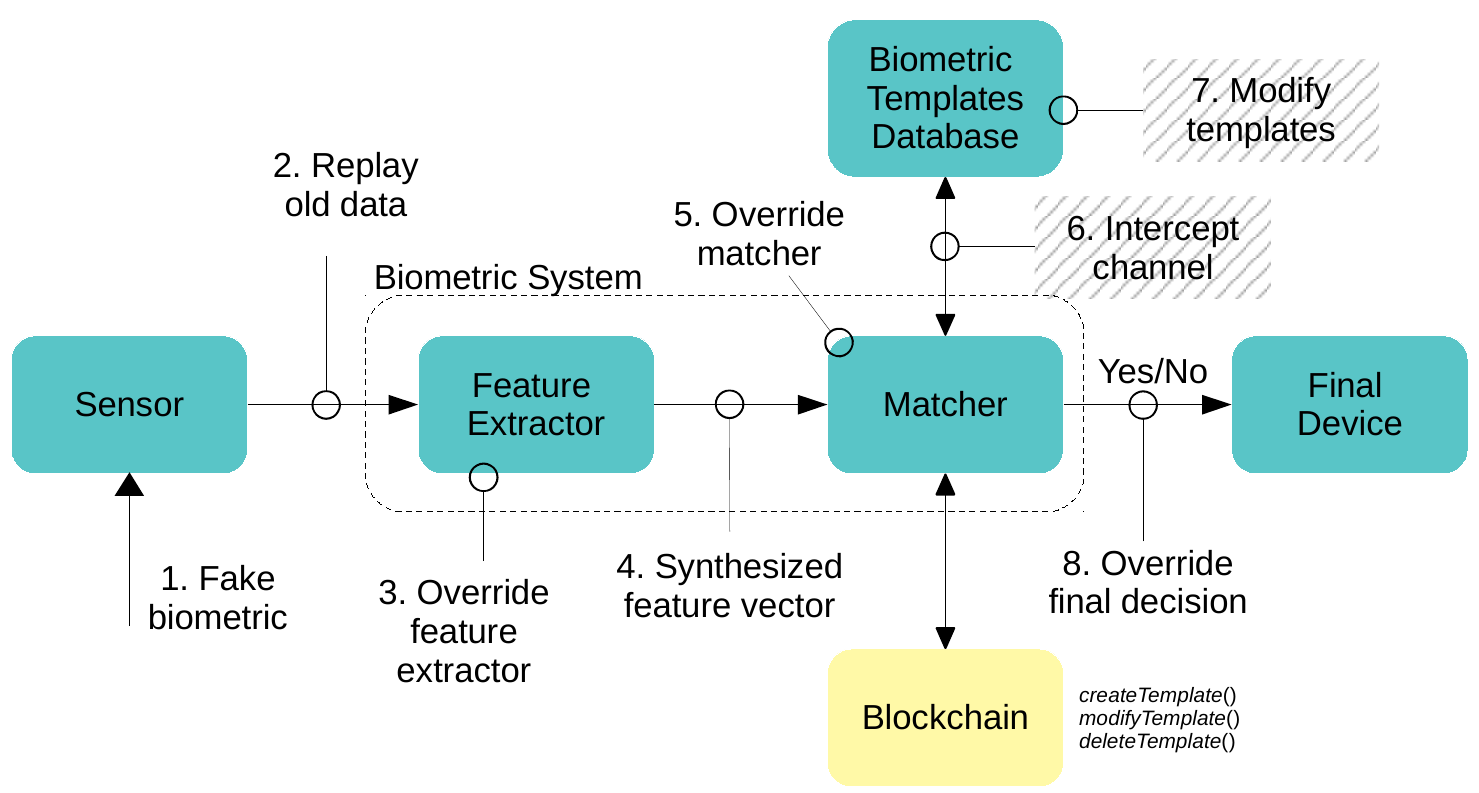}}
\caption{Main modules of a biometric system with indication of its vulnerability points (1 to 8). A public Blockchain is incorporated in exchange of the Biometric Templates Database in order to improve the security of the biometric system. This example integration of biometrics and blockchain is studied experimentally in Section \label{casestudy}.}
\label{fig:proposal}
\end{figure*}

\section{BLOCKCHAIN AND BIOMETRICS}
Once the essential concepts of the blockchain technology have been discussed, the question is: which is the most appropriate architecture for the integration of biometric technology? The short answer is: it depends.

According to the challenges and limitations analyzed in the previous section, one can think that private architectures are clearly the most appropriate. However, public blockchains provide a higher level of security, precisely because of the mining process used in their consensus algorithms. This makes the system more robust against collusion attacks, in which a number of participants manipulate transactions to deceive the rest.

Imagine, for example, a biometric system deployed in a private blockchain. This architecture will force the number of users to be small, since they must be fully identified and trusted. This would limit potential uses, such as a digital identity scheme beyond a company or individual institution. Moreover, if all users must be identified, it is likely that blockchain technology is not actually needed (better solutions may exist).

Therefore, we believe that public blockchain architectures, despite its limitations, will provide safer applications of biometrics of greater social impact. For this reason, the approach studied experimentally in this paper is based on public blockchain architectures.

\subsection{Biometrics and Blockchain Integration}
For integrating biometrics in blockchain systems we can start by analyzing how the usual biometric operations can be performed in a blockchain, so that they cannot be manipulated or falsified. And, if this is possible, which would be its cost and performance.

Undoubtedly, the resulting hybrid biometric system would be more secure. However, this is not an easy goal to achieve because, as stated in Section \ref{sec:privacy}, all the computation performed in a public blockchain is accessible to every participant. This makes necessary to use non-standard encryption or other protection techniques to manage the biometric data in the blockchain. This is specially relevant when incorporating biometrics to blockchain, as biometrics are private data in many cases sensitive. Therefore, as required by best practices or enforced by legislation (e.g., EU GDPR), the biometric data should be stored in a privacy-preserving way (referred to as \emph{protected} in the rest of the paper).

Such advanced encryption schemes include homomorphic encryption and multi-party computation (MPC). However, although some ongoing research is adapting these protection schemes for public blockchains in efficient ways \cite{Zhong2020,Gao2019,Miller2016}, more work to securely integrate biometrics in public blockchains is still needed.

That integration of biometrics and blockchain technologies (public and private) has multiple complex elements and implications in security, storage, and beyond. We therefore foresee that such integration will become a very interesting and active area of research. The work presented in this document is intended as a first step in that direction, especially focused in biometric template protection.

As a first step for advancing such integration we begin taking a look at the different modules of a biometric system, see Fig.~\ref{fig:proposal}:

\begin{itemize}
 \item \textbf{Sensor}: We haven't identified key use cases of the integration of biometric sensors in a blockchain, but this may be possible in the future, e.g., in IoT systems.
 
 \item \textbf{Feature Extractor}: this can be very difficult and expensive to implement in a blockchain, as biometric feature extraction algorithms tend to be complex. Although the variation of this complexity may be large for different types of biometrics, it is likely that the execution costs of running feature extraction in a blockchain will become prohibitive. In any case, this important biometric module would be desirable to be embedded in blockchains for some applications, and further research is needed towards it.
 
 \item \textbf{Biometric Templates Database}: There are different approaches for storing the biometric templates in a blockchain. These approaches are analyzed in Sect. \ref{storage}.
 
 \item \textbf{Matching}: The matching stage can be achieved by simple operations such as distance measurements, although in other cases more complex operations are required (usually, the higher the complexity of the Feature Extractor the simpler the Matcher and vice versa). Therefore, similar comments as discussed before for the Feature Extractor also apply here. In Sect. \ref{sec:face_matching} we fully implement and analyze two matching functions, based in Euclidean and Hamming distances, applied to face biometrics.

\end{itemize}

\begin{figure*}[t]
\centering
\centerline{\includegraphics[width=0.75\textwidth]{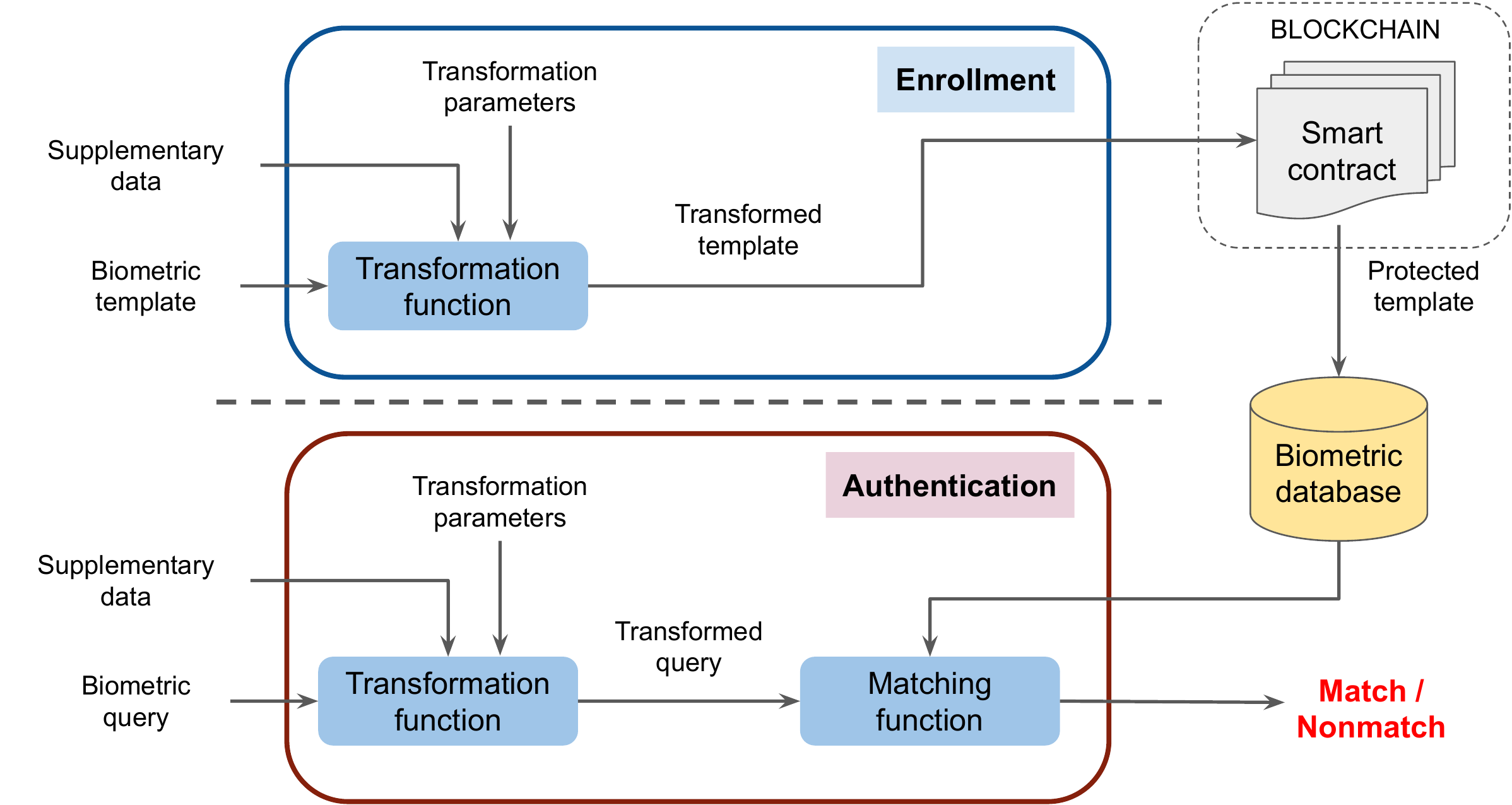}}
\caption{Biometric template protection based on feature transformation (Biometric Hashing).}
\label{fig:TemplateProt}
\end{figure*}

\subsection{Application: Biometric Template Protection}
\label{sec:BTP}
One important application of blockchain to biometrics is \textit{Biometric Template Protection} (BTP). Biometric systems are known to be vulnerable to certain physical \cite{hadid15SPMspoofing} and software attacks \cite{jain_2016}. Physical attacks to the biometric sensor can be overcome to some extent with presentation attack detection (PAD) techniques \cite{2019_Hand_Fierrez}, whereas software attacks can be prevented using biometric template protection techniques. However, the state-of-the-art is still further improvable in many ways \cite{2018_INFFUS_MCSreview2_Fierrez,2017_Access_HEmultiDTW_Marta,2017_PR_multiBtpHE_marta}.

Fig.~\ref{fig:proposal} depicts the typical stages of a biometric system (blue color), including all possible points where a biometric system can be attacked, and our proposed blockchain approach (yellow color) for improving the protection of biometric templates against some of the possible software attacks (gray color). By protecting the traditional template storage with a blockchain, the security level of the resulting biometric system is significantly increased.

Two main attacks are prevented by directly storing the biometric templates in the blockchain, namely: "Modify templates" in the database and "Intercept channel" between database and matcher (points 6 and 7 in Fig.~\ref{fig:proposal}). In most cases, this would not be enough to protect the privacy of the biometric data owners to the largest possible extent.
In order to reach the desirable properties of Irreversibility, Unlinkability and Renewability (IUR) of the biometric templates \cite{2017_PR_multiBtpHE_marta}, the schemes developed so far should be combined in most cases with a more sophisticate Biometric Template Protection (BTP) scheme that enables IUR. BTP techniques can be classified as \cite{2017_PR_multiBtpHE_marta}: 1) Cancelable Biometrics (e.g., Random Projection, BioHashing, BioConvolving, Irreversible Transformation), 2) Biometric Cryptosystems (Fuzzy Vault, Fuzzy Commitment, Secure Sketches), and 3) Biometrics in the Encrypted Domain (Homomorphic Encryption, Garbled Circuits).

\begin{figure*}[t]
     \centering
     \subfigure{\label{fig:data_hashing}
     \includegraphics[width=0.43\linewidth]{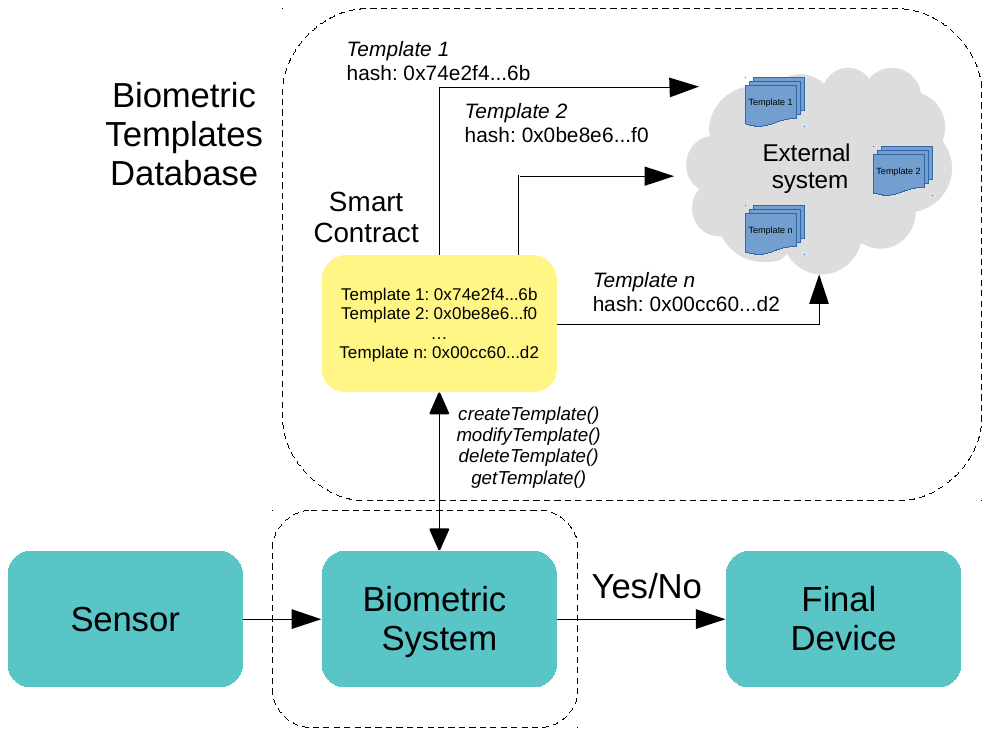}}
     \hspace{0.02\textwidth}
      \subfigure{\label{fig:merkle_tree}
     \includegraphics[width=0.43\linewidth]{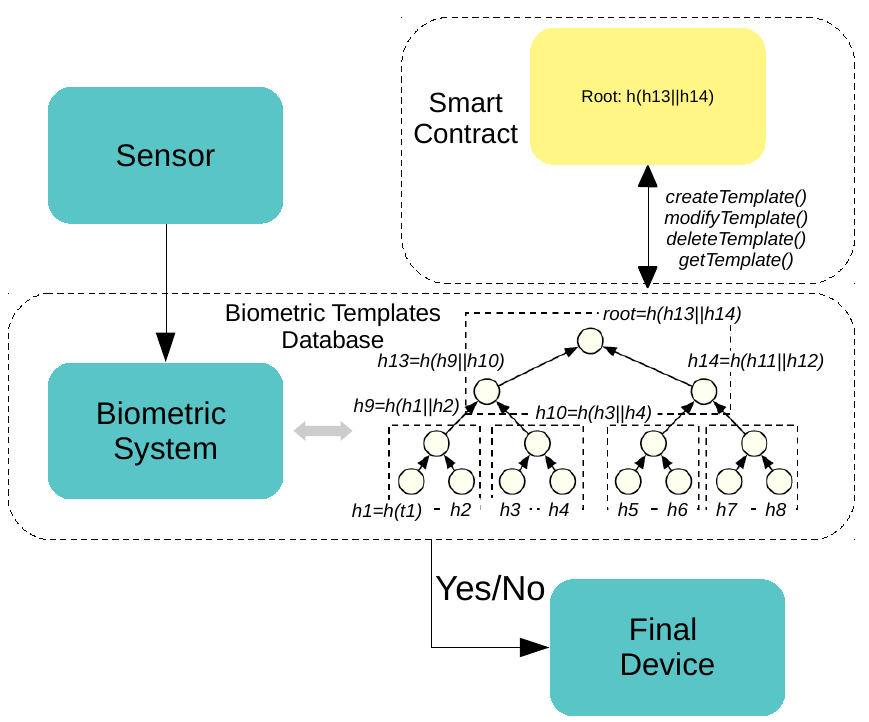}}
     \caption{Biometric systems using data hashing (left) and Merkle trees (right) blockchain storage techniques.}
     \label{fig:hash_merkle_tree}
\end{figure*}

As discussed in \cite{nanda15template}, the strongest methods from the previous BTP classification assume some kind of Supplementary Data (e.g., related to the parameters of the hashing function in Biometric Hashing techniques) entered both at the biometric enrollment and at the biometric matching steps in parallel to the input biometric data (see Fig. \ref{fig:TemplateProt}). For the specific case of BTP based on Biometric Hashing \cite{nanda15template, 2017_PR_multiBtpHE_marta}, the information stored in the blockchain would be a transformation of the input unprotected templates, the input raw biometric data, or a combination of both; and no other changes to the blockchain architectures already discussed would be needed. Given those protected templates, the biometric matching can then be performed in the hashed domain, like in standard cryptographic hashing (see Fig. \ref{fig:TemplateProt}). Note that the matching will be performed similar to a standard cryptographic hashing, but biometric hashing is radically different in nature: biometric hashing aims at providing similar hashes when small changes appear in the input biometric data (smaller than the typical intra-user biometric variability \cite{2011_QualityBio_FAlonso}) and uncorrelated hashes for input changes higher than typical intra-user variability. Adjusting that transition for the specific biometric system at hand is the core of an efficient biometric hashing method.

For more sophisticate BTP methods including Biometric Cryptosystems or Biometrics in the Encrypted Domain, specific adaptations to the blockchain architecture depicted in Fig.~\ref{fig:proposal} may be also required to process elements related to the BTP configuration parameters, keys, or new processing steps included in some BTP methods. This is out of the scope of the present paper and will be investigated in future work.

For the experiments reported in the present paper, we evaluate BTP based on Biometric Hashing for the case of face recognition, which in our experiments is implemented with a fixed-length template representation, following the methodology developed in \cite{Freire2007_ICB}. The other biometric modality considered in the experiments (signature biometrics) is implemented with a state-of-the-art variable-length biometric template \cite{2015_EncyBio_SignFeat}, and this is left out of the BTP experiments, as variable-length representations require further developments beyond the scope of the current paper \cite{2017_Access_HEmultiDTW_Marta}.

\subsection{Storage Schemes}
\label{storage}

In order to discourage its misuse, the cost of storage space in blockchains is specially expensive compared to computation. It is therefore crucial to estimate that cost, in order to minimize it, by selecting the optimal storage scheme. 

There are essentially three main approaches for data storage in blockchains, which are presented below in terms of complexity (from lower to higher), and economic cost (from higher lo lower):

\begin{table}[]
    \centering
        \begin{tabular}{cccc}
        \hline
        \multicolumn{1}{|c|}{\textbf{Operation}} & \multicolumn{1}{c|}{\textbf{Gas/KB}} & \multicolumn{1}{c|}{\textbf{ETH/KB}} & \multicolumn{1}{c|}{\textbf{\$/KB}} \\ \hline
        \multicolumn{1}{|c|}{READ}               & \multicolumn{1}{c|}{6,400}           & \multicolumn{1}{c|}{0.000032}        & \multicolumn{1}{c|}{\$0.005}         \\ \hline
        \multicolumn{1}{|c|}{WRITE}              & \multicolumn{1}{c|}{640,000}         & \multicolumn{1}{c|}{0.0032}          & \multicolumn{1}{c|}{\$0.544}         \\ \hline
        \multicolumn{1}{l}{}                     & \multicolumn{1}{l}{}                 & \multicolumn{1}{l}{}                 & \multicolumn{1}{l}{}
        \end{tabular}
    \caption{Non-volatile storage costs in Ethereum. We have considered a gas price of 1 gwei (1 gwei = $10^{-9}$ ETH), and 1 ETH = \$170 (at time of writing, September 2019).}\label{tab:full-on-chain}
    \end{table}

\begin{itemize}
\item \textbf{Full on-chain storage}: all data is stored, as-is, in the blockchain. For example, biometric templates could be directly stored as a data structure through a smart contract. This is the simplest scheme and therefore, the most inefficient and costly. As an example, Table \ref{tab:full-on-chain} depicts the cost of reading and storing 1 Kilobyte of data in Ethereum in terms of gas units, ether, and US dollars.

\item \textbf{Data hashing}: a more efficient approach is to store the data off-chain and use the blockchain just as an integrity guarantee due to its intrinsic immutability. This way, instead of the full data, only a hash value of it is stored in the blockchain (through smart contracts). Then, the complete template can be stored in any other traditional external storage system (see Fig.~\ref{fig:hash_merkle_tree} (left)).

This scheme provides a great flexibility as any platform can be used to store the full set of biometric templates, e.g., public clouds or existing corporate servers. In any case, to maintain the distributed spirit, resistance to censorship and high availability of public blockchains, distributed storage systems such as IPFS or Swarm \cite{Ozyilmaz2018} would be desirable in this case. In addition, it can make use of any cryptographic hash function, such as the SHA3 family, which can produce outputs from 224 to 512 bits in length. In this study, we consider hashes of 256 bits per template, which, in any case, can greatly reduce storage costs compared to full on-chain storage.

One drawback of this approach is that it is still necessary to ensure the availability of the data stored outside the blockchain. If these data were lost or tampered, even when this modification would be always noticed, the viability of the system would be compromised.

\item \textbf{Linked data structures}: the previous scheme can be further improved through the use of linked data structures, e.g., Merkle trees \cite{Merkle1987}. This specific data structure is widely used in cryptography and computer science problems such as database integrity verification, peer-to-peer networks, and blockchain \cite{Dannen2017}.

A Merkle tree is a binary data structure in which every node contains the cryptographic hash of the concatenation of its child nodes contents. Due to this recursive way of constructing itself, the tree root contains statistical information of the rest of nodes, and the modification of any node content will cause the complete change of the value of the root. This way, the integrity of an arbitrary amount of data can be efficiently assured by arranging this data in a Merkle tree form and securely storing the contents of its root node.

Regarding biometric template protection using blockchains, this structure would store one biometric template at each node, assuring the root node in a smart contract. Therefore, when a new biometric template is created (after the enrollment stage), or an existing one is modified or deleted, the tree is re-calculated and the new root is updated in the blockchain. A simplified scheme of this approach can be found in Fig.~\ref{fig:hash_merkle_tree} (right).

\end{itemize}

\section{EXPERIMENTS}\label{casestudy}

This section describes our experiments integrating biometrics in a blockchain setup. Specifically, we study various practical implementations of \textit{biometric template protection} with \textit{off-chain} and \textit{on-chain matching} in the Ropsten Ethereum testnet. On the one hand, a blockchain-based scheme is added to the traditional template database of two biometric systems (face and signature), with support for the basic operations such as creation, modification, and deletion of the biometric templates. On the other hand, two matching operations (for on-chain matching) are also implemented for the case of face recognition, based on Euclidean and Hamming distances.


Two different biometric traits are considered in this study: \textit{i)} face, and \textit{ii)} handwritten signature. An initial version of this study was presented in~\cite{2019_CVPRw_BlockchainBiometrics_Applications}. The evaluated approaches provide the following advantages:
\begin{itemize}
    \item The modifications to the existing biometric architectures are minimal, so that usual biometric techniques and algorithms (e.g., feature extraction and matching) can be used normally.
    \item For the case of off-chain matching, the evaluated architectures avoid the scalability problems of public blockchains.
    \item No need to use complex smart contracts, which facilitates development and reduces execution costs. Smart contracts do not implement the full biometric processing (e.g., biometric feature extraction is left off-chain), but only the minimum necessary functions to manage the storage of the templates.
\end{itemize}

\subsection{Blockchain Smart Contract}
The biometric templates are generated and accessed through the use of a smart contract, which implements a data structure {\small\texttt{BiometricTemplate}} as a raw array of bytes. This structure is stored in a mapping (or hash table), with an identifier number for the user acting as the mapping key {\small\texttt{mapping(uint => BiometricTemplate)}}. The main operations are described below:
\begin{itemize}
    \item \textbf{Creation:} receives the user ID, template data and metadata, and adds a new {\small\texttt{BiometricTemplate}} structure to the blockchain.
    \item \textbf{Modification:} modifies the template of an existing user. For a hash table storage scheme, this is equivalent to an addition operation.
    \item \textbf{Deletion:} removes the link between a specific template and user ID. Due to the public nature of Ethereum, technically the old template data remains forever in the blockchain.
    \item \textbf{Retrieval:} retrieves the {\small\texttt{BiometricTemplate}} structure for a user. This function is a \textit{call}, not a transaction as the rest of functions. This operation is usually read-only (free to execute), while the previous three operations were potentially state-changing.
\end{itemize}

The smart contract developed can be verified with any blockchain explorer like Etherscan\footnote{0x8f737f448de451db9b1c046be7df3b48839673a1}. It is a basic contract, which does not take care of security issues, and should be used only for experimental purposes.

\subsection{Biometric Systems}
We explore both image-based physiological face biometrics, and signal-based behavioral signature biometrics:

\begin{itemize}
\item \textbf{Face:} we consider the VGG-Face model, one of the most popular face recognition systems based on deep convolutional neural networks (DCNNs)~\cite{vggface}. In this system images are propagated through the CNN obtaining the features at the last fully connected layer. The final matching score is computed through the Euclidean distance of the features obtained from each face image. The dimension of the original face feature vector is 4,096.

\item \textbf{Dynamic Signature:} we consider variable-length templates consisting of a total of 21 time functions, extracted from the normalised \textit{X} and \textit{Y} spatial coordinates across time~\cite{2015_EncyBio_SignFeat}. For the similarity computation, Dynamic Time Warping (DTW) is used to compare the similarity between genuine and query input samples, finding the optimal elastic match among time sequences that minimises a given distance measure.
\end{itemize}

\subsection{Biometric Hashing}
\label{sec:hashing}

As indicated in Section \ref{sec:BTP} we will study experimentally later on Biometric Template Protection based on Biometric Hashing following the approach detailed in \cite{Freire2007_ICB} with a small variation: instead of using genetic programming for selecting the best subset of features, we will apply SFFS feature subset selection (as also discussed but not really applied in \cite{Freire2007_ICB}).

The hashing consists of the concatenation of binary strings extracted from multiple vector-quantized feature subsets with some degree of overlapping features between the different subsets. In the following we describe this hashing process in more detail.

Given a feature vector $\textbf{x}=[x_1,\dots,x_N]$ with $x_i \in
\mathbb{R}$, a biometric hash
\linebreak$\textbf{h}=[h_1,\dots,h_L]$ with $h_i \in \{0,1\}$ of
dimension $L$ is extracted. Let $\textbf{x}^j$ with $j=1,\dots,D$ be
formed by a subset of features of $\textbf{x}$ of dimension $M$
($M<N$), with possibly overlapping features for different $j$. Let
$C^j$ be a codebook obtained by vector quantization of feature
subset $\textbf{x}^j$ using a development set of features
$\textbf{x}^j_{k=1,\dots,K}$. We define $\textbf{h}$ for an input
feature vector $\textbf{x}$ as:

\begin{equation} \label{eq:hash}
\textbf{h}(\textbf{x}) = \mathop{\textrm{concat}}\limits_{j =
1,\dots,D}\{f(\textbf{x}^j,C^j)\}
\end{equation}

\noindent where $f$ is a function that assigns the nearest-neighbour
codewords, and $\textrm{concat}({\cdot})$ denotes the concatenation of
binary strings.

The codebooks $C^j$ are computed with vector quantization as
follows. Let $\textbf{x}^j_{k=1,\dots,K}$ be feature vector subsets
forming a development set. The \emph{k}-means algorithm is used to
compute the centroids of the underlying clusters, for a given number
of clusters $Q$. Then, centroids are ranked based on their distance
to the mean of all centroids. Finally, binary codewords of size
$q=\log_2 Q$ are defined as the position of each centroid in the
ranking using Gray coding.

The threshold $\theta$ represents the maximum number of overlapped features permitted among different subsets, and the best subsets are chosen by using SFFS feature subset selection (minimizing EER in a development dataset).

The parameters of this Biometric Hashing have been tuned heuristically starting from the ones that gave best results in \cite{Freire2007_ICB}, for a final configuration as follows: $N=100$ (size of the real-valued feature vector), $M=4$, and $q=3$.

\subsection{Databases and Experimental Protocol}
Two popular biometric databases are considered for the analysis of face and signature biometrics: Labeled Faces in the Wild (LFW)~\cite{learned2016labeled}, and Biosecure~\cite{ortega2010multiscenario}.

\begin{itemize}
\item \textbf{Face:} experiments are conducted on the LFW database, which is one of the most popular datasets used in face recognition with more than 13,000 face images of famous people collected from the web. We have used the aligned dataset where each image was aligned with funneling techniques.

The VGG-Face pre-trained model was originally trained with VGG-Face database \cite{vggface}, and tested using the unrestricted and outside training data protocols proposed in LFW database \cite{learned2016labeled}. Therefore, there is not extra training for the pre-trained model used here. The evaluation results are computed for 6,000 one-to-one comparisons composed of 3,000 genuine pairs (pairs of images from the same person) and 3,000 impostor pairs (pairs of images belonging to different persons) following the protocols from LFW database.

\item  \textbf{Dynamic Signature:} experiments are conducted on the Biosecure DS2 dataset. This dataset was captured using a Wacom Intuous 3 digitizing tablet with an inking pen in an office-like scenario, providing the following information: \textit{X} and \textit{Y} spatial coordinates, pressure, and timestamp (sampling frequency 100 Hz).

In this study, we consider a set of 50 users. For each user, the first 5 genuine signatures of the first session are used for training, whereas the 15 genuine signatures of the second session are left for testing in order to consider the inter-session variability. Random forgeries are considered as impostors, comparing the training signatures with one genuine signature of the remaining users. The final score is obtained through the average score of the five one-to-one comparisons.

\end{itemize}

\begin{figure*}[t]
     \centering
     \subfigure{\label{fig:global_signature}
     \includegraphics[width=0.47\linewidth]{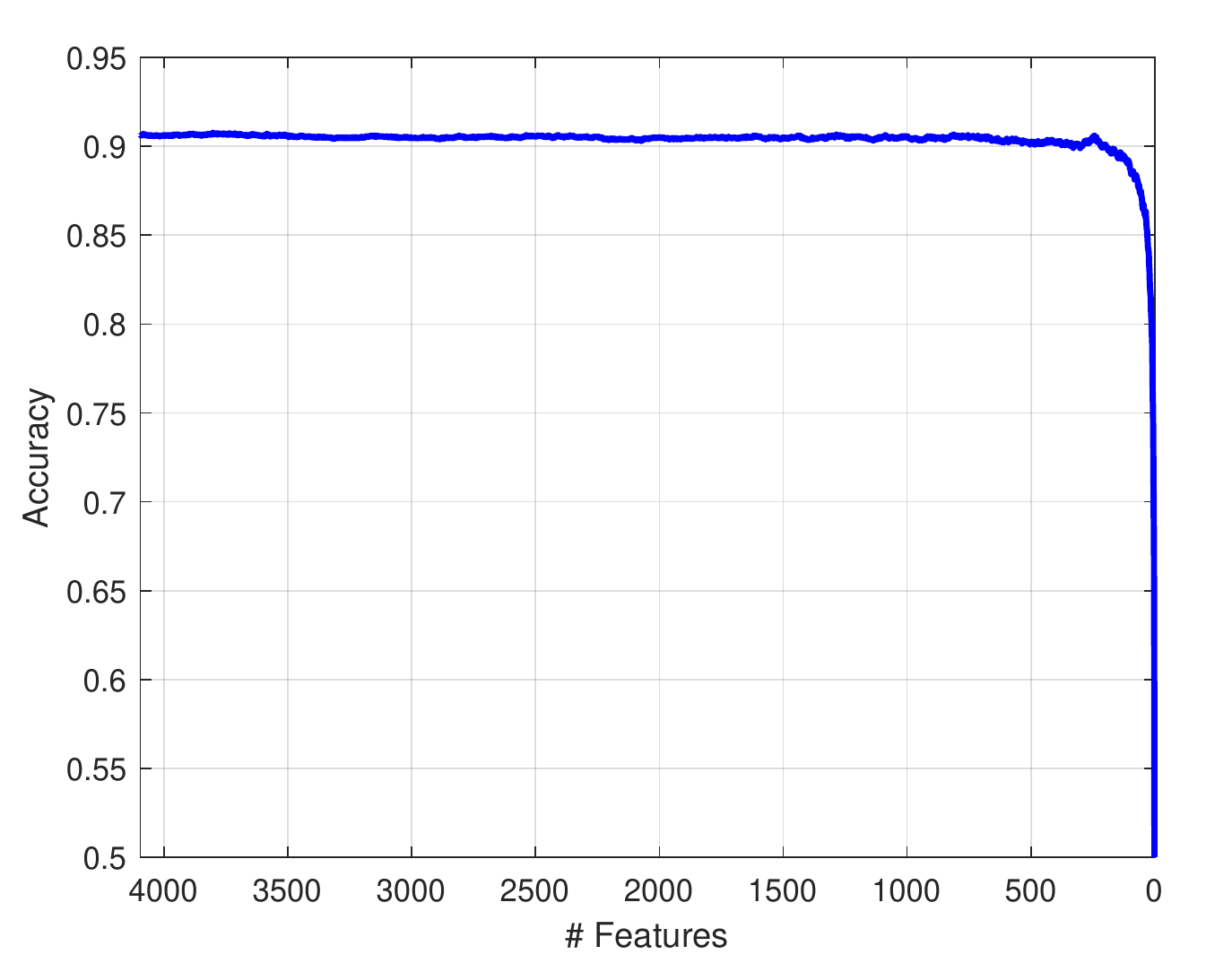}}
      \subfigure{\label{fig:local_signature}
     \includegraphics[width=0.45\linewidth]{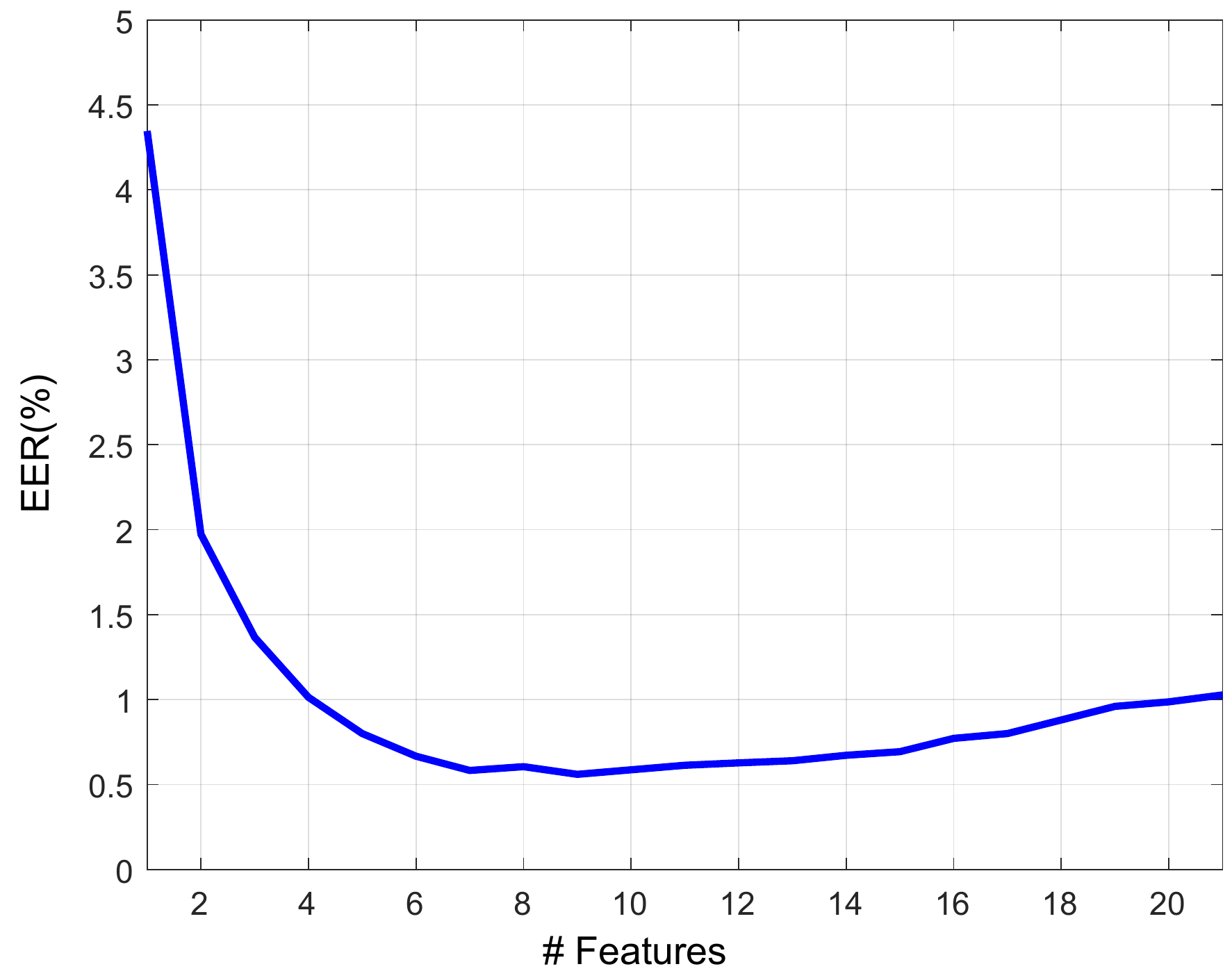}}
     \caption{System performance results in terms of the biometric template size: (left) face, (right) signature, both unprotected cases.}
     \label{fig:signature_size_template}
\end{figure*}

\section{RESULTS}
First, we analyze the system performance of both face and signature biometric systems regarding the size of the biometrics templates. This analysis results crucial in terms of economic cost for the integration of the biometric technology in blockchain. Fig.~\ref{fig:signature_size_template} shows the system performance results in terms of Accuracy and Equal Error Rate (EER) for different sizes of the biometric template. For the case of face biometrics (see Fig.~\ref{fig:signature_size_template} (left)), the system performance is very stable while we gradually remove (randomly) features from the original feature embedding of size 4096. VGG-Face is able to obtain a verification rate with an accuracy (true positive rate) of 89\% using only 100 features (2.5\% from the original 4096 features). This behavior shows that there is a very high redundancy within the feature embedding of the CNN face models, which makes possible to obtain very competitive verification performance while keeping only a small set of features.

\begin{table}[]
\begin{tabular}{|l|c|c|c|}
\hline
\textbf{Case}        & \multicolumn{1}{l|}{\textbf{Overlap ($\theta$)}} & \multicolumn{1}{l|}{\textbf{\# Features}} & \multicolumn{1}{l|}{\textbf{EER (\%)}} \\ \hline
\textbf{Unprotected} & 0                                             & 100 real                              & 11.0                                   \\ \hline
\textbf{Protected}   & 0                                             & 75 binary                                 & 13.6                                   \\ \hline
\textbf{Protected}   & 1                                             & 500  binary                               & 8.5                                    \\ \hline
\textbf{Protected}   & 2                                             & 1500 binary                               & 7.3                                    \\ \hline
\end{tabular}
\caption{Results in terms of EER (\%) for the different approaches for face recognition, from the unprotected case to three protected cases with different feature size (overlap).}\label{tab:BTPresults}
\end{table}

Table \ref{tab:BTPresults} shows the performance results in terms of EER (\%) comparing the standard unprotected case (using Euclidean distance for matching), with three approaches using the biometric template protection scheme already described in Sect. \label{sec:hashing} with three configurations: i) without overlap ($\theta$ = 0) resulting each template in 75 binary numbers, ii) overlap $\theta$ = 1 and then SFFS subset selection fixed to generate 500 binary features for each template, and iii) overlap $\theta$ = 2 also using SFFS to generate 1500 binary features per template. The Hamming distance is applied to the three protected approaches. The results show that the first protected approach ($\theta$ = 0) increases the EER by a relative 24\%. However when the feature overlap increases from $\theta$ = 0 to $\theta$ = 2, the EER decreases a relative 34\% EER going from 11.0\% of the unprotected case to 7.3\% EER. This shows the benefits of using the biometric template approach both in terms of security and system performance.  


For the case of signature biometrics, similar results are obtained improving the EER values when reducing the number of features (see Fig.~\ref{fig:signature_size_template} (right)). The best performance of 0.5\% EER is achieved using a total of 9 local functions (total number of features = 9 local functions $\times$ 343 average time samples per local functions = 3,087 features).

\begin{table*}[]
\centering
\resizebox{\textwidth}{!}{%
\begin{tabular}{|c|c|c|c|c|c|c|}
\hline
\multicolumn{2}{|c|}{\textbf{Biometric}} &
  \multirow{2}{*}{\textbf{Operation}} &
  \multicolumn{3}{c|}{\textbf{Storage scheme}} &
  \textbf{Perfomance} \\ \cline{1-2} \cline{4-7} 
\textit{Scheme} &
  \textit{\begin{tabular}[c]{@{}c@{}}Template\\ size\end{tabular}} &
   &
  \textit{Full on-chain} &
  \textit{\begin{tabular}[c]{@{}c@{}}Data hashing\\ (cost per template)\end{tabular}} &
  \textit{\begin{tabular}[c]{@{}c@{}}Merkle trees\\ (cost for any number \\ of templates)\end{tabular}} &
  \textit{\begin{tabular}[c]{@{}c@{}}Execution time\\ (average)\end{tabular}} \\ \hline
- &
  - &
  \begin{tabular}[c]{@{}c@{}}Smart contract\\ deployment\end{tabular} &
  \multicolumn{3}{c|}{\begin{tabular}[c]{@{}c@{}}498,274 gas\\ (\$0.06972)\end{tabular}} &
  19.19 secs \\ \hline
\multirow{4}{*}{Signature} &
  \multirow{4}{*}{\begin{tabular}[c]{@{}c@{}}Time Functions\\ 3087 x 16 bits\\(unprotected)\end{tabular}} &
  Creation &
  \multirow{2}{*}{\begin{tabular}[c]{@{}c@{}}4,358,990 gas\\ (\$0.741)\end{tabular}} &
  \multicolumn{2}{c|}{\multirow{2}{*}{\begin{tabular}[c]{@{}c@{}}86,848 gas\\ (\$0.0148)\end{tabular}}} &
  12.61 secs \\ \cline{3-3} \cline{7-7} 
 &
   &
  Modification &
   &
  \multicolumn{2}{c|}{} &
  12.85 secs \\ \cline{3-7} 
 &
   &
  Deletion &
  \begin{tabular}[c]{@{}c@{}}504,322 gas\\ (\$0.09)\end{tabular} &
  \multicolumn{2}{c|}{\begin{tabular}[c]{@{}c@{}}18,850 gas\\ (\$0.0032)\end{tabular}} &
   \\ \cline{3-7} 
 &
   &
  Retrieval &
  - &
  - &
  - &
  - \\ \hline
\multirow{16}{*}{Face} &
  \multirow{4}{*}{\begin{tabular}[c]{@{}c@{}}VGG-Face\\ 100 x 32 bits\\ (unprotected)\end{tabular}} &
  Creation &
  \multirow{2}{*}{\begin{tabular}[c]{@{}c@{}}352,912 gas\\ (\$0.06)\end{tabular}} &
  \multicolumn{2}{c|}{\multirow{2}{*}{\begin{tabular}[c]{@{}c@{}}86,848 gas\\ (\$0.0148)\end{tabular}}} &
  \multirow{2}{*}{10.53 secs} \\ \cline{3-3}
 &
   &
  Modification &
   &
  \multicolumn{2}{c|}{} &
   \\ \cline{3-7} 
 &
   &
  Deletion &
  \begin{tabular}[c]{@{}c@{}}49,192 gas\\ (\$0.0083)\end{tabular} &
  \multicolumn{2}{c|}{\begin{tabular}[c]{@{}c@{}}18,850 gas\\ (\$0.0032)\end{tabular}} &
  16.38 secs \\ \cline{3-7} 
 &
   &
  Retrieval &
  - &
  - &
  - &
  - \\ \cline{2-7} 
 &
  \multirow{4}{*}{\begin{tabular}[c]{@{}c@{}}VGG-Face\\ 75 bits\\ (protected)\\ Overlap 0\end{tabular}} &
  Creation &
  \multirow{2}{*}{\begin{tabular}[c]{@{}c@{}}66,544 gas\\ (\$0.01122)\end{tabular}} &
  \multicolumn{2}{c|}{\multirow{12}{*}{N/A}} &
  \multirow{2}{*}{10.53 secs} \\ \cline{3-3}
 &
   &
  Modification &
   &
  \multicolumn{2}{c|}{} &
   \\ \cline{3-4} \cline{7-7} 
 &
   &
  Deletion &
  \begin{tabular}[c]{@{}c@{}}16,173 gas\\ (\$0.00289)\end{tabular} &
  \multicolumn{2}{c|}{} &
  16.38 secs \\ \cline{3-4} \cline{7-7} 
 &
   &
  \begin{tabular}[c]{@{}c@{}}Matching \\ (Hamming distance)\end{tabular} &
  - &
  \multicolumn{2}{c|}{} &
  - \\ \cline{2-4} \cline{7-7} 
 &
  \multirow{4}{*}{\begin{tabular}[c]{@{}c@{}}VGG-Face\\ 500 bits\\ (protected)\\ Overlap 1\end{tabular}} &
  Creation &
  \multirow{2}{*}{\begin{tabular}[c]{@{}c@{}}73,084 gas\\ (\$0.01241)\end{tabular}} &
  \multicolumn{2}{c|}{} &
  \multirow{2}{*}{10.53 secs} \\ \cline{3-3}
 &
   &
  Modification &
   &
  \multicolumn{2}{c|}{} &
   \\ \cline{3-4} \cline{7-7} 
 &
   &
  Deletion &
  \begin{tabular}[c]{@{}c@{}}21,846 gas\\ (\$0.00374)\end{tabular} &
  \multicolumn{2}{c|}{} &
  16.38 secs \\ \cline{3-4} \cline{7-7} 
 &
   &
  Matching &
  - &
  \multicolumn{2}{c|}{} &
  - \\ \cline{2-4} \cline{7-7} 
 &
  \multirow{4}{*}{\begin{tabular}[c]{@{}c@{}}VGG-Face\\ 1500 bits\\ (protected)\\ Overlap 2\end{tabular}} &
  Creation &
  \multirow{2}{*}{\begin{tabular}[c]{@{}c@{}}121,272 gas\\ (\$0.02057)\end{tabular}} &
  \multicolumn{2}{c|}{} &
  \multirow{2}{*}{10.53 secs} \\ \cline{3-3}
 &
   &
  Modification &
   &
  \multicolumn{2}{c|}{} &
   \\ \cline{3-4} \cline{7-7} 
 &
   &
  Deletion &
  \begin{tabular}[c]{@{}c@{}}31,960 gas\\ (\$0.00544)\end{tabular} &
  \multicolumn{2}{c|}{} &
  16.38 secs \\ \cline{3-4} \cline{7-7} 
 &
   &
  Matching &
  - &
  \multicolumn{2}{c|}{} &
  - \\ \hline
\end{tabular}%
}
\caption{Economic costs and performance results for each storage scheme analyzed for unprotected and protected templates. We have considered a gas price of 1 gwei (1 gwei = $10^{-9}$ ETH), and 1 ETH = \$170 (accurate at time of writing, September 2019).}\label{tab:results}
\end{table*}

\subsection{On-chain Face Recognition Matching}\label{sec:face_matching}

In this section, we describe and analyze the on-chain implementation of a usual biometric matching stage based on Euclidean (unprotected case) and Hamming distances (protected cases) for the case of face recognition. 

\subsubsection{Euclidean distance.} The implementation of this operation as a smart contract is specially challenging, due to the lack of native support for floating point arithmetic. To overcome this limitation, we represent decimal numbers as integers (i.e.: 1.12 as 112) and use the Newton-Raphson method to obtain the n\textit{th}-root of \textit{d} by solving the equation $x^n - d = 0$ iteratively.

In this way, we can also adjust the cost of the operation according to the required precision. The source code of the implementation can be found in \cite{GitHub}. The results show that the cost of the calculation of a square root is 23,209 gas units, and 31,304 (around 0.000031 ETH or \$0.00527 for the current ether price) for the calculation of a complete matching operation.

At first glance it may seem a small value, but in a real biometric system, the matching operation must be repeated hundreds or thousands of times. For example, for a scenario with 10,000 users, the total cost for the matching operation (the authentication of each user in the system) would be slightly greater than \$50. Possible solutions and improvements will be be object of future research, as discussed in Section \ref{sec:future_work}.

\subsubsection{Hamming distance.} On the other hand, Hamming distance is significantly easier to implement and, as we will study next, it is virtually free in economic terms.


Every operation implemented in a blockchain must be optimized as much as possible. In this case we use an algorithm whose running time depends on the number of ones present in the binary form of the given number, instead of just traversing and counting the number of ones, obtaining a much better performance.

In contrast to the Euclidean distance, this distance calculation is simpler, and only uses bitwise operations. In addition, it does not store data into the blockchain and, as a result, as a read-only operation has no execution costs. This is specially important in a biometric scenario, where a large number of users can be involved in matching operations. 

\subsection{Analysis of Economic Costs}

Once we have analyzed the size of the biometric templates, we evaluate in Table \ref{tab:results} the costs of the different operations over the templates (creation, modification,  deletion, and retrieval/matching) in units of gas and US dollars, for the biometric technologies and blockchain storage schemes considered.

We first analyze the results for \textit{unprotected} templates, both for signature and face biometrics. These schemes should not be directly used in cases where the privacy of users is to be preserved (as envisioned and enforced, e.g., by the EU GDPR). The unprotected experiments are included mainly for the purpose of setting an initial reference. As expected, the most efficient storage scheme is the one based on Merkle trees. In fact, it is the only one capable of storing any amount of data for the same cost. The rest of the schemes would quickly have a prohibitive cost for the number of templates to be stored in a real environment.

For example, storing a million of templates would cost \$740,000 for the signature system, and \$60,000 for VGG Face using the \textbf{full on-chain} storage scheme. Clearly this is not a realistic option, discouraged not only in economic terms, but also for security and performance reasons.

The \textbf{data hashing} scheme would improve significantly those figures, because it does not store the data itself, but only a hash that guarantees the integrity. For the same scenario, the cost would be a much more reasonable amount of \$14,800 for all the biometric technologies.

Finally, the \textbf{Merkle trees} scheme would imply a cost of only one cent of dollar (\$0.0148) for the storage of any amount of templates. In addition, also the modification operation of a template would have the same cost. However, even for a biometric system operating in a large corporation or environment, these costs seem reasonable.

Of course, all these prices could vary greatly depending on the price of ether, which, as the rest of cryptocurrencies, usually suffers sharp increases and falls in price. However, because it only needs to store 256 bits regardless of the total volume of data, the Merkle tree scheme would still have a reasonable cost in any case.

We now look at the \emph{protected} systems. The storage costs, even for the most expensive scheme (Overlap 2), seem reasonable. For a real scenario with tens of thousands of users, the total cost would range between \$100 and \$200. In addition, the matching function has not associated execution costs, because the Hamming distance instead of Euclidean. The calculation of this distance does not need to permanently store data into the blockchain and, therefore, as a read-only operation, is virtually free. 

Finally, regarding execution time and performance, the experiments show that the proposed system is also viable. It is important to note that this is so even considering that the tests have been conducted in a testnet, where confirmation times are higher and have greater variability than in the main network. Times have been measured performing each operation ten times, discarding the minimum and maximum times, and calculating the average of the rest.

As can be seen, the execution time is slightly higher than 10 seconds for the operations of creation, modification and deletion of templates. These functions are involved only in the user management operations (i.e.: user enrollment), where these delays seems acceptable in usability terms. 

On the other hand, the matching operation, the most demanding in performance terms, can be considered immediate in terms of execution time, because the request is processed by the local Ethereum node, and it does not reach the network.

\section{FUTURE TRENDS}
\label{sec:future_work}

Results shown in section \ref{sec:face_matching} underline the need for additional techniques to implement biometric matching and related operations in blockchain. In addition to the mentioned in previous sections, other possibilities are: i) \textit{zero-knowledge proofs (ZKP)}, and ii) \textit{state channels}, or off-chain computation.

\textbf{ZKPs primitives} allow that one party proves to another, even using a public channel and with presence of an attacker, that they know a secret value $x$, without disclosing the value itself. The process can be even done non-interactive, like zkSNARKs \cite{Fuchsbauer2017}\cite{Parno2013}. Indeed, Ethereum has recently added the arithmetic modular primitives necessary for the  integration of this technology. But, although some recent schemes, as range ZKPs \cite{Morais2019ASO}, could be useful for the threshold checking implicit in the matching operation, all these operations are still very computationally intensive and, therefore, expensive when executed inside a blockchain. 

On the other hand, \textbf{State channels} seems a much more promising technology \cite{Dziembowski2018}. This approach transfers part of the computation outside the chain, so that the involved parties interact directly, without the assistance of a blockchain. Only later, when the exchange of messages and operations is completed, the result is stored and verified by the blockchain.

To date, the most illustrative example of state channels is perhaps \textit{payment channels}, already implemented in popular cryptocurrencies such as Lightning Network for Bitcoin \cite{Seres2019TopologicalAO}. In contrast to traditional payment networks, state channels-based networks allow execution of arbitrary complex smart contracts and, therefore, new possibilities beyond the mere transfer of money.

In biometric systems, state channels could be applied to the matching operation. This way, the modules in charge of the user authentication would operate as usual, but digitally signing each performed step. At the end of the process, this approach would provide a fully certified chain of operations in the blockchain. As only the result (i.e. user successfully authenticated or not) is stored, the process would be much cheaper both in computational and economic terms.

The exact way in which this scheme could be implemented must be object of further research.

\section{CONCLUSION}

In this study we have discussed the opportunities and challenges in the integration of blockchain and biometrics, with emphasis in biometric template storage and protection, a key problem in biometrics still largely unsolved. Key tradeoffs involved in that integration, namely, latency, processing time, economic cost, and biometric performance, are experimentally studied through the implementation of a smart contract on the Ethereum blockchain platform, with related code made public in github.\footnote{\url{https://github.com/BiDAlab/BlockchainBiometrics}}

We have first discussed the main storage schemes for public blockchains (Ethereum), and implemented a smart contract for the estimation of its storage cost. The results obtained have proved that straightforward schemes such as the direct storage of the biometric templates on-chain, or direct data hashing, are not appropriate for real biometric systems. Nevertheless, when liked data structures such as Merkle trees are used, the storage cost has become fixed regardless of the total volume of data, reducing execution times (between 10 - 20 seconds for writing operations). Reading operations (retrieving) of templates are usually free of cost and very fast to execute, because they are locally processed.

We have also explored both qualitatively and quantitatively the main performance and cost factors of: i) off-chain and on-chain biometric matching, and ii) storing the biometric templates with or without template protection. On-chain matching has been demonstrated to be feasible and cost-effective for simple matchers based on Hamming distance. Additionally, we have shown that adequate template protection can preserve the privacy of the templates (as recommended by best practices in privacy-preserving biometrics and related legislation such as EU GDPR), and at the same time it can maintain and even improve the biometric matching accuracy with respect to unprotected templates.

Finally, we have discussed future trends in blockchain technologies that may be beneficial for a deeper, more efficient, and more productive integration of blockchain and biometrics, with special attention to zero-knowledge proofs (and variants) and state-channels.

\section{ACKNOWLEDGMENT}
This work has been supported by projects: PRIMA (H2020-MSCA-ITN-2019-860315), TRESPASS-ETN (H2020-MSCA-ITN-2019-860813), BIBECA (MINECO/FEDER RTI2018-101248-B-I00), COPCIS (MINECO/FEDER TIN2017-84844-C2-1-R), Bio-Guard (Ayudas Fundaci\'on BBVA 2017), and Cecabank. Ruben Tolosana is supported by Consejer\'ia de Educaci\'on, Juventud y Deporte de la Comunidad de Madrid y Fondo Social Europeo.

\bibliographystyle{IEEEtran}
\bibliography{biblio}

\begin{IEEEbiography}{Oscar Delgado-Mohatar}{\,}received the B.S.degree in Computer Science from the Universidad Politecnica in 2002, and the PhD in Telecommunications Engineering by Universidad Carlos III de Madrid in 2011. His research interests include cryptology, network security, cryptocurrencies and blockchain. He leads the chair on blockchain technologies at Universidad Autonoma de Madrid funded by Grant Thornton.
\end{IEEEbiography}

\begin{IEEEbiography}{Julian Fierrez}{\,}received the M.Sc. and Ph.D. degrees in telecommunications engineering from the Universidad Politecnica de Madrid, Spain, in 2001 and 2006, respectively. Since 2004 he has been at Universidad Autonoma de Madrid, where he is currently an Associate Professor. From 2007 to 2009 he was a Visiting Researcher with Michigan State University, USA, under a Marie Curie Fellowship. His research interests include signal and image processing, pattern recognition, and biometrics, with an emphasis on multibiometrics, biometric evaluation, system security, forensics, and mobile applications of biometrics. He has been actively involved in multiple EU projects focused on biometrics (e.g., TABULA RASA and BEAT), and has attracted notable impact for his research. He was a recipient of a number of distinctions, including the EAB Industry Award 2006, the EURASIP Best Ph.D. Award 2012, and the 2017 IAPR Young Biometrics Investigator Award. He is Associate Editor of the IEEE TRANSACTIONS ON INFORMATION FORENSICS AND SECURITY and the IEEE TRANSACTIONS ON IMAGE PROCESSING.
\end{IEEEbiography}

\begin{IEEEbiography}{Ruben Tolosana}{\,}received the M.Sc. degree in Telecommunication Engineering, and his Ph.D. degree in Computer and Telecommunication Engineering, from Universidad Autonoma de Madrid, in 2014 and 2019, respectively. In April 2014, he joined the Biometrics and Data Pattern Analytics - BiDA Lab at the Universidad Autonoma de Madrid, where he is currently collaborating as a PostDoctoral researcher. Since then, Ruben has been granted with several awards such as the FPU research fellowship from Spanish MECD (2015), and the European Biometrics Industry Award (2018). His research interests are mainly focused on signal and image processing, pattern recognition, and machine learning, particularly in the areas of face manipulation, human-computer interaction and biometrics. He is author of several publications and also collaborates as a reviewer in many different high-impact conferences (e.g., ICDAR, IJCB, ICB, BTAS, EUSIPCO, etc.) and journals (e.g., IEEE TPAMI, TCYB, TIFS, TIP, ACM CSUR, etc.). Finally, he has participated in several National and European projects focused on the deployment of biometric security through the world.
\end{IEEEbiography}

\begin{IEEEbiography}{Ruben Vera-Rodriguez}{\,}received the M.Sc. degree in telecommunications engineering from Universidad de Sevilla, Spain, in 2006, and the Ph.D. degree in electrical and electronic engineering from Swansea University, U.K., in 2010. Since 2010, he has been affiliated with the Biometric Recognition Group, Universidad Autonoma de Madrid, Spain, where he is currently an Associate Professor since 2018. His research interests include signal and image processing, pattern recognition, and biometrics, with emphasis on signature, face, gait verification and forensic applications of biometrics. He is actively involved in several National and European projects focused on biometrics. Ruben has been Program Chair for the IEEE 51st International Carnahan Conference on Security and Technology (ICCST) in 2017; and the 23rd Iberoamerican Congress on Pattern Recognition (CIARP 2018) in 2018.
\end{IEEEbiography}

\end{document}